\documentclass[12pt]{article}
\usepackage[utf8]{inputenc}
\usepackage[T1]{fontenc}
\usepackage{lmodern}
\usepackage{geometry}
\usepackage{sectsty}
\usepackage{enumitem}
\usepackage{amsmath}
\usepackage{amsfonts}
\usepackage{amssymb}
\usepackage{parskip}
\usepackage{xcolor}
\usepackage{booktabs}
\usepackage{tabularx}
\usepackage{caption}
\usepackage{graphicx}
\usepackage{hyperref}
\usepackage{CJKutf8}
\usepackage{url}
\usepackage{authblk}
\sectionfont{\large\bfseries}
\subsectionfont{\normalsize\bfseries}

\begin{document}
	
	\title{The fragility of ``cultural tendencies'' in LLMs}
	\author[1,2]{\footnotesize Kun Sun\thanks{\texttt{email: kunsun@tongji.edu.cn}}}
\author[2,3]{\footnotesize Rong Wang\thanks{\texttt{email: rong.wang@ims.uni-stuttgart.de}}}

\affil[1]{\footnotesize Department of Linguistics, Tongji University, Shanghai, China}
\affil[2]{\footnotesize Department of Computational Linguistics, Tübingen University, Germany}
\affil[3]{\footnotesize The Institute of Natural Language Processing, Stuttgart University, Stuttgart, Germany}
	\date{}
	\maketitle

In a recent study, Lu, Song, and Zhang \cite{lu2025cultural} (LSZ) propose that large language models (LLMs), when prompted in different languages, display culturally specific tendencies. They report that the two models ( i.e., \texttt{gpt} and \texttt{ernie}) respond in more interdependent and holistic ways when prompted in Chinese, and more independent and analytic ways when prompted in English. LSZ attribute these differences to deep-seated cultural patterns in the models, claiming that prompt language alone can induce substantial cultural shifts. While we acknowledge the empirical patterns they observed, we find their experiments, methods, and interpretations problematic. 
In this paper, we critically re-evaluate the methodology, theoretical framing, and conclusions of LSZ. We argue that the reported ``cultural tendencies'' are not stable traits but fragile artifacts of specific models and task design. To test this, we conducted targeted replications using a broader set of LLMs and a larger number of test items. Our results show that prompt language has minimal effect on outputs, challenging LSZ’s claim that these models encode grounded cultural beliefs. 

 While LSZ’s data show output variation across two languages, their interpretation that LLMs possess culture-like tendencies is not warranted by their methodology or experiments. They present interesting response patterns, but the limited number of test items and narrow model sample undermine the validity and generalizability of their conclusions.

First, \textit{language is not assumed to be neutral}, and the conceptual novelty is limited. LSZ claim to challenge cultural neutrality in prompts but cite no work endorsing such neutrality. In NLP and AI, it is well established that language shapes model behavior via training data, tokenization, and linguistic norms \cite{bender2018data, blodgett2020language, bolukbasi2016man, caliskan2017semantics}. Their study therefore documents a plausible but underexplored effect rather than overturning a consensus. Prior work \cite{adilazuarda2024towards, demszky2023using, shanahan2023role} has already shown that LLM outputs vary across languages and prompt frames. LSZ’s main contribution is isolating language as the manipulation variable and linking it to cultural psychology scales.

Second, three major methodological concerns further weaken the study. The first concern is ``anthropomorphism''. The study applies human psychometric metrics (e.g., collectivism scale, attribution bias task) to LLMs. While LLMs can produce text aligned with cultural constructs, interpreting this as genuine social orientation or cognitive style risks conflating statistical mimicry with real cognition \cite{mitchell2021ai, pellert2024ai}. 
Second, causal attribution is unclear, since the effects could stem from training data imbalances, linguistic structures, or tokenization differences rather than cultural cognition \cite{liu2025cultural, gaebler2024auditing}. Third, there is a lack of transparency, as the authors withheld test items, mitigation details, and prompt templates, which prevents verification and overlooks open-science norms, despite the well-known sensitivity of LLM outputs to prompt design \cite{he2024does,sahoo2024systematic}. The details on these issues are seen in Section B of the \textbf{Supplementary Information}.

Finally, the testing scope is too narrow for robust conclusions: only 4–12 items per task across seven tasks, and just two models (\texttt{gpt} and \texttt{ernie}) were evaluated. Such small item counts reduce statistical power, making reliable or valid inferences difficult. Even if accurate, the results would at best indicate model-specific differences, not generalizable culture-like tendencies across LLMs.

 To reassess the validity and robustness of LSZ’s findings, we conducted four targeted experiments (with an extra experiment) using the same vignettes and measures as their seven tasks. We retained the original response formats (e.g., remain Likert ratings, forced-choice) but expanded both model coverage and test size. Specifically, we evaluated eight leading LLMs in both English and Chinese, including \texttt{gpt} and \texttt{ernie} as controls. All models were state-of-the-art versions: \texttt{gpt-4.1}, \texttt{gemini-2.5}, \texttt{claude-sonnet-4}, \texttt{grok-4}, \texttt{kimi-k2}, \texttt{deepseek-V3}, \texttt{ernie-4.5}, and \texttt{qwen3-30b}. Each experiment included between 30 and 60 items in English or Chinese, with prompts matched across languages.
 
 Because LSZ did not release their test items and prompt templates, exact replication was impossible. Instead, we constructed effective replications by adapting cultural psychology scale items cited in LSZ to match their described approach while covering a broader range of topics. Our experiments are summarized in Table~\ref{tab:our-experiments}. Unlike LSZ, we required models to justify their responses for each test item, encouraging more deliberate reasoning. All experiments were run via LLM APIs, using two prompt templates: one with a real-world situational role (task + role + instruction + format + examples) and one without.  All test items were executed in batch mode with \texttt{temperature = 0} to maximize determinism, and each model completed all conditions in both English and Chinese. We replicated LSZ’s statistical analyses using various \texttt{t}-tests and also applied additional methods including ANOVA, effect size estimation, item-level analysis, and logistic regression to improve comparability and rigor. Full details of each experiment and the differences from LSZ’s design are provided in Section C of the \textbf{Supplementary Information}. The complete datasets, prompts, resulting data, and  programming scripts are available at \url{https://osf.io/hv2g8/}.

\begin{table}[!ht]
	\centering
	\caption{Summary of our replicated experiments. (Note: All experiments were designed as the replications of LSZ, also using newly created stimuli that matched the original constructs while expanding model coverage and item counts)}
	\scalebox{0.75}{
		\begin{tabular}{p{0.28\linewidth} p{0.3\linewidth} p{0.38\linewidth} p{0.23\linewidth}}
			\toprule
			\textbf{Experiment} & \textbf{LSZ's Hypothesis} & \textbf{Methodology} & \textbf{\# Test items per language} \\
			\midrule
			\textbf{Exp 1: Cultural Stability and Entity Change} & Chinese prompts evoke a stronger ``change-framed'' orientation (belief in malleability) than English prompts. & 1) English and Chinese prompts with change-framed and neutral versions. 2) Topics cover three similar social orientation scales from LSZ: collectivism scale, individual–collective primacy scale. 3) Seven-point Likert ratings with reasons for ratings. 4)Replicates the first three similar tasks of  Social orientation in LSZ & 60 items; 4 scores per item ( vs. 21 items in LSZ) \\
			\midrule
			\textbf{Exp 2: Cultural Norms and Attribution Bias} & Chinese prompts lead to greater situational attribution than English prompts. & 1) Scenarios in English and Chinese, instructing to respond as an ``average Chinese person'' (dispositional/internal traits vs. situational/external pressures). 2) Seven-point Likert ratings with reasons. 3) Replicates attribution bias task in LSZ. & 30 vignettes; four scores per vignettes ( vs. 12 vignettes in LSZ)  \\
			\midrule
			\textbf{Exp 3: Cultural Logic and Reasoning} & Chinese cultural framing increases acceptance of conclusions aligned with collectivist values, even if formally invalid. & 1) Logical syllogisms (collectivist-consistent vs. inconsistent) in English and Chinese. 2) Binary judgments of intuitive acceptability (yes/no; \begin{CJK}{UTF8}{gbsn}是/否\end{CJK}) with reasons. 3) Replicates intuitive vs. formal reasoning task in LSZ. & 30 items; one choice per item ( vs. 4 items in LSZ) \\
			\midrule
			\textbf{Exp 4: Expectations in Interpersonal Relationships} & Predicting possible changes in interpersonal relationships. & 1) Expectations about relationship development. 2) Seven-point Likert ratings with justifications(reasons). 3) Replicates expectation of change task in LSZ. & 30 items; one score per item ( vs. 4 items in LSZ) \\
			\midrule
			\textbf{Additional Exp: Picture Choice} & Inclusion of other in the self scale. & Uses the exact visual stimulus from LSZ and replicates their forth task. & 4 items; one choice \\
			\bottomrule
		\end{tabular}
	}
	\label{tab:our-experiments}
\end{table}

After completing all experiments, the main results are summarized in Table~\ref{tab:result} (see also Table~8 and Fig.~1 in the \textbf{Supplementary Information}). The two API prompt templates produced no significant differences. For comparability with LSZ, we report only the results from the template with situational scenarios.

\begin{table}[!ht]
	\centering
	\small
	\caption{The main results of the experiments (EN = English; ZH = Chinese; ``origin'' indicates ZH vs. EN models. The \textit{p}-value is considered significant when it is smaller than 0.05.)}
	\scalebox{0.75}{
		\begin{tabular}{p{2.8cm} p{4.5cm} p{3.8cm} p{3.2cm} p{3.8cm}}
			\toprule
			\textbf{Experiment} & \textbf{ZH vs. EN LLMs} & \textbf{\texttt{gpt} vs. \texttt{ernie}} & \textbf{Prompt language effect} & \textbf{Significant models} \\
			\midrule
			\textbf{1. Cultural stability} &
			Small framing effect ($\textit{p} < 0.05$); no origin/language effect ($\textit{p}=0.15$). &
			Similar performance, no language difference ($\textit{p}>0.4$). &
			language effect negligible (no overall effect ). &
			\texttt{deepseek} and \texttt{qwen} prefer ZH ($0.01<\textit{p}<0.05$), no effects in the other six models \\
			\midrule
			\textbf{2. Attribution bias} &
			No origin/language effect overall ($\textit{p}=0.25$); minor language sensitivity in some models. &
		Language preference for either \texttt{gpt} or \texttt{ernie}. & 
			 language effect minimal (no overall effect). &
			\texttt{ernie}, \texttt{gpt} and \texttt{qwen} senitive to EN ($\textit{p}<0.01$). \\
			\midrule
			\textbf{3. Syllogism quality} &
			No origin/language effect ($\textit{p}=1$); knowledge tasks easier ($\textit{p}=0.025$) &
			Comparable \texttt{ernie} and \texttt{gpt} performance; no language sensitivity. &
			No language effect. & No significant models (Chi-Square test, logistic regression and more tests, etc.)
			 \\
			\midrule
			\textbf{4. Interpersonal expectations} &
			No origin/language effect ($\textit{p}=0.917$); language sensitivity in some models &
			\texttt{gpt} better with Chinese prompts ($\textit{p}=0.017$); \texttt{ernie} no difference. &
			No overall language effect. &
			Only \texttt{kimi} and \texttt{gpt} show significant language effects ($0.01<\textit{p}<0.05$). \\
			\bottomrule
	\end{tabular}}
	\label{tab:result}
\end{table}

As shown in Table~\ref{tab:result}, we found negligible effects of language or model origin across the experiments. Prompt language influenced only a few models on specific tasks, with \texttt{gpt} showing sensitivity in two tasks and \texttt{ernie} in one, offering limited support for the original claim. The other three English LLMs showed no significant effects in any experiment. In the additional picture task, only \texttt{ernie} differed by prompt language, consistent with LSZ’s results, but this sensitivity was absent in the remaining five of seven tasks.

\begin{figure}[!ht]
	\centering
	\includegraphics[ width=\textwidth]{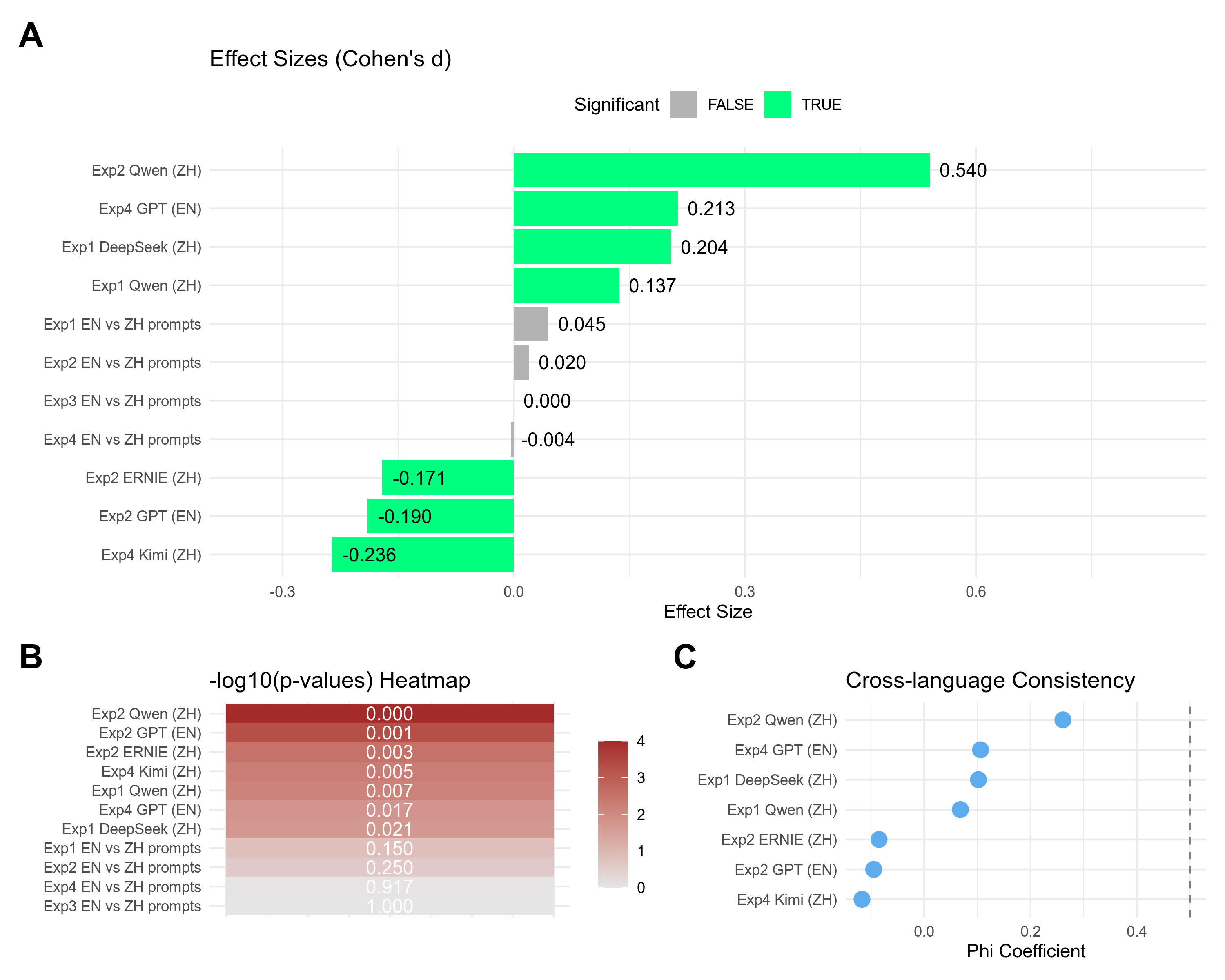}
	\caption{
		Results across experiments, visualized in three panels. 
		\textbf{(A)} Effect sizes (Cohen's $d$) for each comparison. Positive values indicate higher scores for the first condition; negative values indicate higher scores for the second condition. Green bars denote statistically significant differences ($p < 0.05$), while grey bars denote non-significant results. 
		\textbf{(B)} Heatmap of statistical significance expressed as $-\log_{10}(p)$. Darker brown corresponds to smaller $p$-values (greater significance). 
		\textbf{(C)} Cross-language consistency, measured by the Phi coefficient ($\phi$), for the signifcant models in all Experiments. Values close to~1 indicate high agreement between choices under different prompt languages. 
	}
	\label{fig:vis_plot}
\end{figure}

The main findings in these experiments are summarized in Table~\ref{tab:combined}. The effect sizes, heatmaps, and Phi coefficients corresponding to the main results are shown in Figure~\ref{fig:vis_plot}. Note that Cohen's $d$ can be converted to Pearson's $r$ using the formula
\[
r = \frac{d}{\sqrt{d^2 + 4}}.
\]
In the special case of a $2 \times 2$ contingency table, the Phi coefficient ($\phi$) is equivalent to Pearson's $r$, so we have $	\phi = r$. This provides a straightforward way to translate effect sizes between these commonly used metrics.

However, no model displayed consistent cultural tendencies across varying prompts, tasks, or content conditions. Although isolated task-specific effects appeared, such as minor shifts in individual experiments, these effects were unstable across tasks and did not generalize across models or domains. For example, effects observed for some models (e.g., \texttt{gpt},\texttt{kimi}, and \texttt{ernie}) lacked consistency, undermining claims of stable cultural tendencies. This instability challenges the idea that such behaviors reflect intrinsic cultural traits rather than temporary, task-dependent, and model-specific responses.

In summary, our experiments found no prompt-language variations that consistently produced language effects, indicating that prompt language alone is not a reliable driver of cultural orientation in LLMs. The results suggest that cultural and linguistic influences in these tasks are minimal, unstable, or absent. We conclude that ``cultural tendencies'' reported in LSZ show little generalizability in mainstream LLMs. Detailed results and analyses are provided in Section D of the \textbf{Supplementary Information}.

Finally, we believe that the cultural patterns observed by LSZ are more plausibly explained by task-specific artifacts rather than stable, culturally grounded cognition within LLMs. This interpretation carries important implications for related research domains. From a policy standpoint, evaluation protocols should clearly distinguish between correlations arising from linguistic form or cultural framing and claims regarding embedded cultural cognition. Such differentiation is critical to avoid anthropomorphizing statistical models and to ensure that governance frameworks address substantive risks rather than methodological artifacts. Overinterpreting LLM outputs as evidence of cultural cognition risks misinforming AI governance, particularly in culturally sensitive contexts where the statistical limitations of these systems remain unacknowledged. In sum, developing robust, transparent, and comparative methodologies is crucial for establishing a credible evidentiary basis for the cultural and cognitive behaviors attributed to LLMs. \footnote{We reported these issues to \textit{Nature Human Behaviour} (NHB). However, the journal responded that ``we do not believe that it provides further clarification around the original study, nor does it fundamentally challenge or criticize its conclusions to the extent we look for''. Therefore, we are sharing our short paper here to bring these concerns to the community’s attention. A recent comment titled ``How to evaluate the cognitive abilities of LLMs'' \cite{ivanova2025evaluate}, published in NHB, outlines principles for assessing the cognitive abilities of LLMs. Unfortunately, the paper by LSZ, along with several recent studies on LLMs’ abilities on NHB, clearly contradicts both the principles proposed in that comment and the evaluation criteria commonly adopted in the AI research community.}

\bibliographystyle{plain}

\bibliography{reference}

\begin{thebibliography}{10}

\bibitem{adilazuarda2024towards}
Muhammad~Farid Adilazuarda, Sagnik Mukherjee, Pradhyumna Lavania, Siddhant
  Singh, Alham~Fikri Aji, Jacki O'Neill, Ashutosh Modi, and Monojit Choudhury.
\newblock Towards measuring and modeling" culture" in llms: A survey.
\newblock {\em arXiv preprint arXiv:2403.15412}, 2024.

\bibitem{aron1992inclusion}
Arthur Aron, Elaine~N Aron, and Danny Smollan.
\newblock Inclusion of other in the self scale and the structure of
  interpersonal closeness.
\newblock {\em Journal of personality and social psychology}, 63(4):596, 1992.

\bibitem{bender2018data}
Emily~M Bender and Batya Friedman.
\newblock Data statements for natural language processing: Toward mitigating
  system bias and enabling better science.
\newblock {\em Transactions of the Association for Computational Linguistics},
  6:587--604, 2018.

\bibitem{blodgett2020language}
Su~Lin Blodgett, Solon Barocas, Hal Daum{\'e}~III, and Hanna Wallach.
\newblock Language (technology) is power: A critical survey of “bias” in
  nlp.
\newblock In {\em Proceedings of the 58th Annual Meeting of the Association for
  Computational Linguistics}, pages 5454--5476, 2020.

\bibitem{bolukbasi2016man}
Tolga Bolukbasi, Kai-Wei Chang, James~Y Zou, Venkatesh Saligrama, and Adam~T
  Kalai.
\newblock Man is to computer programmer as woman is to homemaker? debiasing
  word embeddings.
\newblock {\em Advances in neural information processing systems}, 29, 2016.

\bibitem{caliskan2017semantics}
Aylin Caliskan, Joanna~J Bryson, and Arvind Narayanan.
\newblock Semantics derived automatically from language corpora contain
  human-like biases.
\newblock {\em Science}, 356(6334):183--186, 2017.

\bibitem{chang2024survey}
Yupeng Chang, Xu~Wang, Jindong Wang, Yuan Wu, Linyi Yang, Kaijie Zhu, Hao Chen,
  Xiaoyuan Yi, Cunxiang Wang, Yidong Wang, et~al.
\newblock A survey on evaluation of large language models.
\newblock {\em ACM transactions on intelligent systems and technology},
  15(3):1--45, 2024.

\bibitem{Chen1998toward}
Y.-R. Chen, J.~Brockner, and T.~Katz.
\newblock Toward an explanation of cultural differences in in-group favoritism:
  the role of individual versus collective primacy.
\newblock {\em Journal of Personality and Social Psychology}, 75:1490--1502,
  1998.

\bibitem{demszky2023using}
Dorottya Demszky, Diyi Yang, David~S Yeager, Christopher~J Bryan, Margarett
  Clapper, Susannah Chandhok, Johannes~C Eichstaedt, Cameron Hecht, Jeremy
  Jamieson, Meghann Johnson, et~al.
\newblock Using large language models in psychology.
\newblock {\em Nature Reviews Psychology}, 2(11):688--701, 2023.

\bibitem{derksen2022kinds}
Maarten Derksen and Jill Morawski.
\newblock Kinds of replication: Examining the meanings of “conceptual
  replication” and “direct replication”.
\newblock {\em Perspectives on Psychological Science}, 17(5):1490--1505, 2022.

\bibitem{gaebler2024auditing}
Johann~D Gaebler, Sharad Goel, Aziz Huq, and Prasanna Tambe.
\newblock Auditing large language models for race \& gender disparities:
  Implications for artificial intelligence-based hiring.
\newblock {\em Behavioral Science \& Policy}, 10(2):46--55, 2024.

\bibitem{gallegos2024bias}
Isabel~O Gallegos, Ryan~A Rossi, Joe Barrow, Md~Mehrab Tanjim, Sungchul Kim,
  Franck Dernoncourt, Tong Yu, Ruiyi Zhang, and Nesreen~K Ahmed.
\newblock Bias and fairness in large language models: A survey.
\newblock {\em Computational Linguistics}, 50(3):1097--1179, 2024.

\bibitem{he2024does}
Jia He, Mukund Rungta, David Koleczek, Arshdeep Sekhon, Franklin~X Wang, and
  Sadid Hasan.
\newblock Does prompt formatting have any impact on llm performance?
\newblock {\em arXiv preprint arXiv:2411.10541}, 2024.

\bibitem{hu2025generative}
Tiancheng Hu, Yara Kyrychenko, Steve Rathje, Nigel Collier, Sander van~der
  Linden, and Jon Roozenbeek.
\newblock Generative language models exhibit social identity biases.
\newblock {\em Nature Computational Science}, 5(1):65--75, 2025.

\bibitem{hui1988measurement}
C~Harry Hui.
\newblock Measurement of individualism-collectivism.
\newblock {\em Journal of Research in Personality}, 22(1):17--36, 1988.

\bibitem{ivanova2025evaluate}
Anna~A Ivanova.
\newblock How to evaluate the cognitive abilities of llms.
\newblock {\em Nature Human Behaviour}, 9(2):230--233, 2025.

\bibitem{Ji2001cul}
L.-J. Ji, R.~E. Nisbett, and Y.~Su.
\newblock Culture, change, and prediction.
\newblock {\em Psychological Science}, 12:450--456, 2001.

\bibitem{Kitayama2006vo}
S.~Kitayama, K.~Ishii, T.~Imada, K.~Takemura, and J.~Ramaswamy.
\newblock Voluntary settlement and the spirit of independence: evidence from
  japan’s “northern frontier”.
\newblock {\em Journal of Personality and Social Psychology}, 91:369--384,
  2006.

\bibitem{liang2022holistic}
Percy Liang, Rishi Bommasani, Tony Lee, Dimitris Tsipras, Dilara Soylu,
  Michihiro Yasunaga, Yian Zhang, Deepak Narayanan, Yuhuai Wu, Ananya Kumar,
  et~al.
\newblock Holistic evaluation of language models.
\newblock {\em arXiv preprint arXiv:2211.09110}, 2022.

\bibitem{lin2021truthfulqa}
Stephanie Lin, Jacob Hilton, and Owain Evans.
\newblock Truthfulqa: Measuring how models mimic human falsehoods.
\newblock {\em arXiv preprint arXiv:2109.07958}, 2021.

\bibitem{liu2025cultural}
Zhaoming Liu.
\newblock Cultural bias in large language models: A comprehensive analysis and
  mitigation strategies.
\newblock {\em Journal of Transcultural Communication}, 3(2):224--244, 2025.

\bibitem{lu2025cultural}
Jackson~G Lu, Lesley~Luyang Song, and Lu~Doris Zhang.
\newblock Cultural tendencies in generative ai.
\newblock {\em Nature Human Behaviour}, pages 1--10, 2025.

\bibitem{masuda2001attending}
Takahiko Masuda and Richard~E Nisbett.
\newblock Attending holistically versus analytically: comparing the context
  sensitivity of japanese and americans.
\newblock {\em Journal of Personality and Social Psychology}, 81(5):922, 2001.

\bibitem{meyerowitz1987effect}
Beth~E Meyerowitz and Shelly Chaiken.
\newblock The effect of message framing on breast self-examination attitudes,
  intentions, and behavior.
\newblock {\em Journal of Personality and Social Psychology}, 52(3):500, 1987.

\bibitem{mitchell2021ai}
Melanie Mitchell.
\newblock Why ai is harder than we think.
\newblock In {\em Proceedings of the Genetic and Evolutionary Computation
  Conference}, pages 3--3, 2021.

\bibitem{morris1994culture}
Michael~W Morris and Kaiping Peng.
\newblock Culture and cause: American and chinese attributions for social and
  physical events.
\newblock {\em Journal of Personality and Social psychology}, 67(6):949, 1994.

\bibitem{Norenzayan2002cul}
A.~Norenzayan, E.~E. Smith, B.~J. Kim, and R.~E. Nisbett.
\newblock Cultural preferences for formal versus intuitive reasoning.
\newblock {\em Cognitive Science}, 26:653--684, 2002.

\bibitem{nosek2022replicability}
Brian~A Nosek, Tom~E Hardwicke, Hannah Moshontz, Aur{\'e}lien Allard,
  Katherine~S Corker, Anna Dreber, Fiona Fidler, Joe Hilgard, Melissa
  Kline~Struhl, Mich{\`e}le~B Nuijten, et~al.
\newblock Replicability, robustness, and reproducibility in psychological
  science.
\newblock {\em Annual Review of Psychology}, 73(1):719--748, 2022.

\bibitem{pellert2024ai}
Max Pellert, Clemens~M Lechner, Claudia Wagner, Beatrice Rammstedt, and Markus
  Strohmaier.
\newblock Ai psychometrics: Assessing the psychological profiles of large
  language models through psychometric inventories.
\newblock {\em Perspectives on Psychological Science}, 19(5):808--826, 2024.

\bibitem{sahoo2024systematic}
Pranab Sahoo, Ayush~Kumar Singh, Sriparna Saha, Vinija Jain, Samrat Mondal, and
  Aman Chadha.
\newblock A systematic survey of prompt engineering in large language models:
  Techniques and applications.
\newblock {\em arXiv preprint arXiv:2402.07927}, 2024.

\bibitem{shanahan2023role}
Murray Shanahan, Kyle McDonell, and Laria Reynolds.
\newblock Role play with large language models.
\newblock {\em Nature}, 623(7987):493--498, 2023.

\bibitem{tversky1981framing}
Amos Tversky and Daniel Kahneman.
\newblock The framing of decisions and the psychology of choice.
\newblock {\em Science}, 211(4481):453--458, 1981.

\bibitem{wei2022chain}
Jason Wei, Xuezhi Wang, Dale Schuurmans, Maarten Bosma, Fei Xia, Ed~Chi, Quoc~V
  Le, Denny Zhou, et~al.
\newblock Chain-of-thought prompting elicits reasoning in large language
  models.
\newblock {\em Advances in neural information processing systems},
  35:24824--24837, 2022.

\bibitem{Yamaguchi1994c}
S.~Yamaguchi.
\newblock {\em Collectivism among the Japanese: a perspective from the self},
  pages 175--188.
\newblock Sage Publications, 1994.

\bibitem{Yoo2011m}
B.~Yoo, N.~Donthu, and T.~Lenartowicz.
\newblock Measuring hofstede’s five dimensions of cultural values at the
  individual level: development and validation of cvscale.
\newblock {\em Journal of International Consumer Marketing}, 23:193--210, 2011.

\end{thebibliography}

\newpage

\appendix

\textbf{The Supplementary Information}




\section{Introduction}

In a recent study, Li, Song, and Zhang \cite{lu2025cultural} (LSZ) argue that generative AI models such as \texttt{gpt} and \texttt{ernie} exhibit stable cultural tendencies when used in different languages, that is, more interdependent and holistic in Chinese, and more independent and analytic in English. Although their empirical findings are statistically robust, we argue that their study over-interprets these outputs as indicative of culturally grounded cognition. We show that these behaviors are more plausibly explained as surface-level and temporary statistical mimicry in specific tasks rather than stable, internalized cultural orientation. 
To support our claims, we introduce a set of replicated experiments demonstrating the instability and content-driven nature of these effects. We advocate for a less anthropomorphic and more mechanistic interpretation of LLMs' (large language moodels) behavior in cross-cultural settings.

\section{Critique of Research and Methods}

LSZ conducted several experiments to collect empirical evidence on output variation between the two languages. However, the study appears to face notable theoretical and methodological limitations, which we outline below.

First, \textit{language is not assumed to be neutral}. This was explained in the main text. The second concern is \textit{the problem of anthropomorphism}. Crucially, the study re-purposes human-targeted psychometric metrics (e.g., the collectivism scale, attribution bias task) to probe LLMs with no beliefs, introspection, or socialized cognition. While models can output text statistically aligned with cultural constructs, interpreting this as ``social orientation'' or ``cognitive style'' might anthropomorphize fundamentally statistical systems (i.e., anthropomorphism). For example, a collectivism scale assumes a respondent’s self-concept, but LLMs lack a `self', generating responses basically based on token probabilities. The tendency to assign cultural psychology constructs to LLM outputs potentially risks conflating mimicry with cognition. The mimicry of human-like conversation can easily be mistaken for genuine understanding or cognition \cite{pellert2024ai}. 

Third, \textit{prompt framing confounds interpretation}. The fact that cultural tendencies can be altered by a simple framing instruction (e.g., ``Assume the role of a Chinese person'') undermines the claim of intrinsic tendencies. The observed effects are better interpreted as context-sensitive pattern completion, not stable cultural encodings \cite{lin2021truthfulqa, wei2022chain}. LLMs tend to follow textual cues, not cultural norms.

Forth, \textit{causal attribution remains unclear}. The study attributes observed behaviors to culture embedded in language, yet does not disentangle whether effects arise from cultural semantics, linguistic structure, or surface-level patterns in training data \cite{liu2025cultural}. This leaves causal mechanisms ambiguous. Without auditing the models' training corpora or controlling for translation artifacts, the claim that LLMs ``exhibit cultural tendencies'' remains speculative. For instance, the observed effects may stem from training data imbalances (e.g., more collectivist-themed texts in Chinese corpora) or tokenization differences between languages, rather than cultural cognition.

Fifth, a major methodological flaw is the \textit{lack of transparency about the test items} and the claimed ``mitigation efforts'' against LLM memorization. The authors state they used items from established scales but did not release any prompts or stimuli, even in sample form. Many such items (very limited test items appeared in the cited published papers) are publicly available and likely in LLM training data. Without the exact materials, one cannot tell whether results reflect genuine cultural tendencies or simple recall. Only one of seven tasks (attribution bias) used nonpublic data. The authors provided no details on their mitigation methods, rendering their findings unverifiable and contradicting the growing call for data transparency and open science in AI. Moreover, prompt format and structure strongly influence LLM outputs \cite{sahoo2024systematic}, but the authors neither shared their prompt templates nor explained their design in detail, making their AI evaluation work difficult to assess.

Additionally, the conclusions of LSZ face considerable challenges regarding validation and generalizability due to the limited scope of their tests. The number of test items is relatively small, typically ranging from 4 to 12 across seven tasks. Second, only two LLMs (\texttt{gpt} and \texttt{ernie}) were evaluated. Even if correct, the findings merely show that \texttt{gpt} and \texttt{ernie} differ in their responses to social culture, not that English and Chinese LLMs differ, as two models alone lack sufficient diversity. Third, the cultural psychology aspects examined are narrowly focused on social orientation and cognitive style, despite cultural psychology encompassing a much broader range of factors. This limited number of test items, restricted model variety, and narrow psychological scope collectively constrain the robustness and generalizability of their findings \cite{chang2024survey,   gallegos2024bias, liang2022holistic, hu2025generative}. To strengthen their conclusions, additional statistical methods should be applied to cross-validate the results.

\section{Experimental Design to Re-evaluate ``Cultural Tendencies''}

\subsection{General design}
To test the validity and robustness of the original findings, we conducted three targeted experiments using the same vignettes and measures as LSZ, but with significantly broader model coverage and a larger item pool. Specifically, we included eight leading LLMs in both English and Chinese to increase diversity and robustness: \texttt{openai/gpt-4.1}, \texttt{google/gemini-2.5-flash}, \texttt{anthropic/claude-sonnet-4}, \texttt{x-ai/grok-4}, 
\texttt{kimi-k2},\\ 
\texttt{deepseek/deepseek-V3}, \texttt{baidu/ernie-4.5-300b-a47b}, and \texttt{qwen/qwen3-30b-a3b}. Each experiment was built around a 30--60-item test set in either English or Chinese (i.e., equivalent English and Chinese prompts) , which could address the original study’s limited scope. 

Note that LSZ did not release their stimuli or prompts, preventing exact replication. Our test items therefore aim for a conceptual replication, closely mirroring the original study. Based on related work and sources cited by LSZ, we adapted existing cultural psychology scale items to cover diverse topics while matching their described approach. The experiments we designed are summarized in Table~1 in the main text. All experiments were done using LLMs' API to make implementations. All eight LLMs (four English and four Chinese)  were tested in both Chinese and English test items in each experiment. 

\subsection{Replications}

Our experimental design aimed to closely replicate the core constructs examined by LSZ, while substantially broadening the scope in terms of model diversity, item quantity, and linguistic coverage \cite{ derksen2022kinds,  nosek2022replicability}. Since the original study did not publicly release their stimuli or exact prompt formulations, we undertook a careful process to reconstruct test items that reflect the psychological constructs central to their work. We drew upon cultural psychology literature and related validated scales cited by LSZ, adapting them into scenarios and tasks suitable for LLMs.

\textbf{Experiment 1} examined cultural stability and implicit theories of change, building on established principles from social and cross-cultural psychology. We developed 60 social scenarios, each presented in two versions. One depicted a neutral social situation, while the other, a change-framed version, explicitly highlighted human capacity for growth and adaptation over time. This constitutes a classic framing manipulation, designed to prime either a fixed (entity) or a malleable (incremental) mindset.

Building on the test item examples and the cited work in LSZ \cite{Chen1998toward,Yamaguchi1994c,Yoo2011m}, our approach mirrors LSZ in employing key social orientation measures, including the collectivism scale, the individual cultural values–collectivism scale, and the individual–collective primacy scale. The three tasks in LSZ are highly similar in topic, testing content, and implementation; therefore, we merged them into a single integrated task for our experiment.
Whereas LSZ employed direct survey prompts, we embedded these constructs within richer scenario-based items, measured on a continuous 0–1 scale with seven gradations. This provided finer resolution than traditional categorical formats and enabled experimental testing of how information framing influences beliefs about personal and social change \cite{ hui1988measurement, tversky1981framing,  meyerowitz1987effect}.

In LSZ's study, these scales were presented directly as survey items (not embedded in scenarios), with responses measured on Likert-type scales. In contrast, our Experiment 1 measures the same constructs, but instead of direct questions, we used scenario-based framing to experimentally manipulate mindsets. Our first experiment is a better replication of the social orientation component of LSZ, though it adds a causal framing test that the original did not. In this way, Experiment 1 serves as a better replication of LSZ’s assessment of social orientation, extending it by integrating framing manipulations to test causal effects on change-related attitudes.


In \textbf{Experiment 2}, we investigated cultural norms and attribution by providing models with 30 everyday behavioral scenarios. All our item tests refer to the example test items in LSZ and the examples of their cited work \cite{Kitayama2006vo}. We also refer to the other related work \cite{ masuda2001attending, morris1994culture}. Ultimately, we collected 30 vignettes, each accompanied by four statements, with participants providing individual ratings for each statement on a seven-point Likert scale along with justifications. This approach successfully replicated the attribution bias task reported in LSZ.


\textbf{Experiment 3} targeted cultural logic and reasoning by presenting 30 syllogistic arguments varying in their alignment with collectivist values by referring to the example provided by LSZ and their cited work \cite{Norenzayan2002cul}. The models were asked to assess whether the conclusions logically followed from the premises, responding ``Yes'' or ``No'' (\begin{CJK}{UTF8}{gbsn}是/否\end{CJK}` in Chinese prompt). This task captured the tension between formal logical validity and relational or cultural appropriateness, reflecting how collectivist reasoning may override strict formal logic in favor of harmony and social context. This replication mirrors the intuitive versus formal reasoning tasks in the original study.

\textbf{Experiment 4} is to evaluate how language models assess and justify expectations about the development of interpersonal relationships when presented with controlled, context-specific scenarios. This approach enables comparison of models’ sensitivity to explicit social cues, their tendencies toward neutral or extreme judgments, and their ability to provide consistent, evidence-based reasoning. The broader goal is to examine how factors such as model architecture, training data, and prompt language influence relational prediction patterns.

Drawing on the example test item in LSZ and its cited work \cite{Ji2001cul}, we developed a set of 30 items. Each item presents a brief scenario involving two individuals in a defined social context, followed by a single predictive statement about their future relationship. Models evaluate the statement on a 1–7 likelihood scale and provide a 20–50 word explanation based solely on scenario details. All items are administered under controlled conditions to minimize bias, and scoring incorporates both quantitative ratings and qualitative adherence to the specified reasoning constraints.
Additionally, as for the additional picture task, we used the three test item (one from the original study, and the other three from \cite{aron1992inclusion}. 



Although the original stimuli were not publicly released, we carefully selected and adapted these scale items to cover a diverse range of topics, following the methodological procedures described by LSZ. By doing so, we ensured that our test items aligned closely with the theoretical constructs and task designs of the original study while allowing for an expanded and rigorous empirical investigation. This approach allowed us to maintain the integrity of the original research questions and constructs while overcoming the limitations posed by the absence of original materials.

Despite this, unlike the original study, which employed relatively small and unspecified item sets, our experiments applied substantially larger pools of test items (i.e., ranging from 30 to 60 per experiment) to improve reliability and representativeness. Furthermore, while the original study evaluated only two LLMs, we broadened the scope to eight state-of-the-art large language models spanning both English- and Chinese-language backgrounds. Response formats from the original tasks, including Likert ratings, forced-choice classification, and binary logical judgments, were preserved to facilitate meaningful comparison of outcomes (see Table~\ref{tab:method-contrast}). 

\subsection{Prompt template}

To implement the experiments via API calls to various LLMs, we designed two complementary prompt templates that differed in their inclusion of situational context. The first template explicitly incorporated a real-world role and background, for example instructing the model to adopt the persona of ``You will participate in a psychology experiment about entity change, assessing the likelihood of improvement in various situations''.  This was followed by clear task instructions, specification of the input/output format, and illustrative examples to ensure clarity and guide the model’s responses, as shown in Table~\ref{tab:prompt_structure}. This situational framing was intended to activate culturally grounded reasoning patterns and align the model’s outputs with the targeted social context.

\vspace{-0.5em}
\begin{table}
	\centering
	\caption{Summary of Prompt Structure}
	\scalebox{0.8}{
		\begin{tabular}{|l|p{12cm}|}
			\hline
			\textbf{Item} & \textbf{Example} \\
			\hline
			Task & Entity Change Cultural Experiment \\
			\hline
			Situation & You will participate in a psychology experiment about entity change, assessing the likelihood of improvement in various situations. Please read the provided scenario and rate the probability of improvement with a value between 0 and 1 based on your judgment. \\
			\hline
			Instruction & 1) Read the provided scenario carefully. Rate each of the four statements in the exact order provided, using the following scale:
			- 1 = Strongly disagree
			- 2 = Disagree
			- 3 = Slightly disagree
			- 4 = Neutral
			- 5 = Slightly agree
			- 6 = Agree
			- 7 = Strongly agree
			2) Base your ratings solely on the information in the scenario. Do not make assumptions about unstated details (e.g., cultural background, personality traits, or environmental conditions).
			3) If the scenario lacks sufficient detail to evaluate a statement, use a neutral rating (4) and explain the lack of evidence in your reasoning.
			4) Provide a concise reasoning paragraph (20–50 words) that justifies each rating by referencing specific scenario details and explaining the relative influence of personal versus environmental factors.
			\\
			\hline
			Format & Input format: \{ ``id'': ``<ID>'', ``scenario'': ``<Scenario Description>'', ``version'': ``<change\_framed or neutral>'' \}, Output format: \{ ``id'': ``<ID>'', ``scenario'': ``<Scenario Description>'', ``version'': ``<change\_framed or neutral>'', ``score'': ``<Value between 0 and 1>'', ``reasoning'': ``<Brief explanation for choosing this probability>'' \} \\
			\hline
			Example & Input: \{ ``id'': ``EC01-A'', ``scenario'': ``Two people are in conflict over a disagreement. People often grow, adapt, and change as time passes. How likely is it that their relationship will improve over time?'', ``version'': ``change\_framed'' \}, Output: \{ ``id'': ``EC01-A'', ``scenario'': ``Two people are in conflict over a disagreement. People often grow, adapt, and change as time passes. How likely is it that their relationship will improve over time?", ``version'': ``change\_framed'', ``score'': 0.7, ``reasoning'': ``The explicit mention that people grow and change suggests improvement is likely over time.'' \} \\
			\hline
	\end{tabular}}
	
	\label{tab:prompt_structure}
\end{table}
\vspace{-0.5em}

The second template removed explicit situational role information, presenting only task instructions, input/output formats, and examples without any cultural framing or persona induction. This allowed us to isolate the effect of explicit cultural role prompting on model behavior and to test whether the mere presence of culturally relevant instructions influenced outcomes.

All eight models were evaluated under both prompt conditions in both English and Chinese languages, resulting in a fully crossed design that explored interactions between model origin, prompt framing, and linguistic context.

\subsubsection{Implementations}

All experiments were conducted in a fully automated and controlled manner using batch processing with the model temperature parameter fixed at 0 to ensure deterministic outputs. This setting minimized random variability across repeated runs and enabled precise evaluation of model responses to our test stimuli. Each of the eight large language models: four primarily US-based (\texttt{gpt-4.1}, \texttt{gemini-2.5}, \texttt{claude-sonnet-4}, \texttt{grok-4}) and four primarily China-based (\texttt{kimi-k2}, \texttt{deepseek-V3}, \texttt{ernie-4.5}, \texttt{qwen3-30b}), was tested via their respective APIs \footnote{US-based models = English models = English models; China-based models = Chinese models}. All models completed every test item in both Chinese and English across all experimental conditions, ensuring comprehensive cross-linguistic and cross-model coverage.

The test scripts, prompt files, and evaluation pipelines were standardized across all experiments and models, ensuring consistency and reproducibility. This implementation strategy allowed us to efficiently generate large volumes of data and perform robust statistical analyses. Additionally, all code, data, and prompt templates have been made publicly available (available at: \url{https://osf.io/hv2g8/} ) to facilitate transparency and future research replication.

\subsection{Statistical tests}

In order to maintain comparability with the original work, we initially employed two-sided \texttt{t}-tests analogous to those reported by LSZ. However, to strengthen the methodological robustness of our analyses, we also incorporated a broader suite of statistical procedures. Analysis of variance (ANOVA) models were applied to investigate potential interactions between language of the prompt, model origin, and prompt framing conditions. Effect size metrics were computed to evaluate the practical significance of observed differences beyond mere statistical significance.

At the item level, we conducted detailed analyses to detect patterns of response variability across individual test items, which helped reveal unique trends that may be obscured in aggregate scores. Furthermore, categorical grouping analyses were performed to explore how model responses distributed across theoretically meaningful classes, such as dispositional versus situational attributions or logically valid versus invalid syllogisms. This comprehensive analytic approach provided richer insights into the stability and replicability of cultural tendencies in LLMs. Note that due to the binary trait in Experiment 3, we also added Chi-Square test, McNemar's test, logistic regression, Phi Coefficients etc. to enhance the test validations. 

\subsection{The differences between our experiments and LSZ}


The main differences are summarized in Table~\ref{tab:method-contrast}. The test datasets, prompts, the final results, and the programming scripts in the three experiments are available at: \url{https://osf.io/hv2g8/}.

\vspace{-0.5em}
\begin{table}
	\centering
	\caption{Methodological comparison between LSZ and our replicated experiments. (Note: Our replication was conceptual rather than literal due to lack of released materials, but followed the same constructs while expanding scope.)}
	\scalebox{0.8}{
		\begin{tabular}{p{0.28\linewidth} p{0.3\linewidth} p{0.3\linewidth}}
			\toprule
			\textbf{Dimension} & \textbf{LSZ} & \textbf{Our Replicated Experiments} \\
			\midrule
			\textbf{Models tested} & 2 models (\texttt{gpt-4-1106-preview}, \texttt{ernie-3.5-8K-0205}) & 4 US-based models (\texttt{gpt-4.1}, \texttt{gemini-2.5}, \texttt{claude-sonnet-4}, \texttt{grok-4}), and China-based models (\texttt{kimi-k2}, \texttt{deepseek-V3}, \texttt{ernie-4.5}, \texttt{qwen3-30b}) \\ \hline
			\textbf{Language conditions} & Chinese vs. English (translation) & Chinese vs. English (equivalent) \\ \hline
			\textbf{Test items} & Unclear  & Used the explicit test items and added more from being adopted from the existing research mentioned in LSZ ; with diverse topics \\ \hline
			\textbf{Prompt style} & General instructions, no explicit role prompts; \texttt{temperature=0} & Two prompt templates: (1) with situation role + instruction + format + examples, (2) without situational role but the others remain; \texttt{temperature=0} \\ \hline
			\textbf{Constructs measured} & Social orientation (4 tasks), cognitive style (3 tasks) & Same constructs mapped to all tasks in LSZ but add much more test items covering diverse topics \\ \hline
			\textbf{Number of items per task} & 4--12 items & 30--60 items per condition (30--60 total per language) \\ \hline
			\textbf{Model origin coverage} & 1 US-based model, 1 China-based model & 4 US-based models, 4 China-based models \\ \hline
			\textbf{Number of items per task} & 4--12 items & 30--60 items per condition (30--60 total per language) \\ \hline
			\textbf{Response reason} & No requirements & Explicitly requiring models to provide justifications for their responses, encouraging more deliberate and thoughtful outputs \\ \hline
			\textbf{Statistical methods} & Two-sided $t$-tests + Chi-square test & $t$-tests + Wilcoxon tests + ANOVA + effect sizes + item-level and logic regression etc.\\ \hline
			\textbf{Replication type} & Original stimuli (unpublished for some tasks) & Replication + Conceptual replication: replicate all original tasks, and newly constructed stimuli matching task structure and constructs, due to lack of released materials \\
			\bottomrule
	\end{tabular}}
	\label{tab:method-contrast}
\end{table}
\vspace{-0.5em}

\section{Results}

\subsection{Experiment 1}
\noindent
\textbf{Experiment~1} assessed the effects of prompt language (English vs.\ Chinese), LLM origin (Chinese vs.\ English), and framing (neutral vs.\ change-framed) on model performance to rate for the tasks of cultural stability and entity change across 960 observations, using paired and independent $t$-tests, Wilcoxon tests, and a three-way ANOVA.

The results are summarized in Table~\ref{tab:exp1results}. A smaller but statistically significant main effect was observed for \textbf{LLM origin}: Chinese LLMs outperformed English LLMs by $0.021$ points overall ($p = 0.0089$, $d = 0.169$, $\eta^{2} = 0.0071$). This advantage was most evident for Chinese-prompt inputs (difference $= 0.0283$, $p = 0.0144$), while the English-prompt condition showed no significant origin effect ($p = 0.219$), suggesting a mild cultural--linguistic alignment benefit.

By contrast, \textbf{prompt language} showed no statistically detectable influence on performance overall (mean difference $ = 0.0053$, $p = 0.1501$, $d = 0.045$, $\eta^{2} = 0.0002$), and this null effect held within both Chinese LLMs ($p = 0.0210$ with a very small effect) and English LLMs ($p = 0.4432$). While two models (DeepSeek, Qwen) showed significant language preferences, the effect was inconsistent in direction and negligible in magnitude for the others, indicating largely language-agnostic performance at the group level.

Additionally, the other statistical effects are summarized in Table~\ref{tab:exp1results1}. As shown in Table~\ref{tab:exp1results1} the clearest and largest effect was \textbf{framing} (vs. neutral). Change-framed prompts yielded markedly higher scores than neutral prompts (mean difference $ = -0.1650$, $p < 2.2 \times 10^{-16}$, Cohen’s $d = -2.095$), an effect confirmed as the dominant source of variance in the ANOVA ($\eta^{2} = 0.5423$), explaining more than half of the total performance variability. This framing advantage was highly robust, appearing across all statistical approaches and for all eight models tested. the ANOVA detected a significant interaction between LLM origin and framing ($p = 1.28 \times 10^{-5}$, $\eta^{2} = 0.0083$): both Chinese and English LLMs benefited from change-framing, but the extent of the boost differed by origin, pointing to possible differences in framing sensitivity across model families.


\vspace{-0.5em}
\begin{table}
	\centering
	\caption{The main results for Experiment~1. (Model-specific language effects (Chinese vs. English prompts)}
	\scalebox{0.8}{
		\begin{tabular}{l l r r r c r l l}
			\hline
			\textbf{Model} & \textbf{Origin} & \textbf{Mean Diff.} & \textbf{$t$} & \textbf{$p$} & \textbf{Sig.} & \textbf{$d$} & \textbf{Effect Mag.} & \textbf{Direction} \\
			\hline
			\texttt{deepseek} & Chinese & $0.025$ & $2.374$ & $\textbf{0.021}$ & * & $0.204$ & Small & \textbf{Chinese higher} \\
			\texttt{qwen} & Chinese & $0.015$ & $2.801$ & $\textbf{0.007}$ & ** & $0.137$ & Negligible & \textbf{Chinese higher} \\
			\texttt{kimi} & Chinese & $-0.008$ & $-1.236$ & $0.221$ &  & $-0.076$ & Negligible & No diff. \\
			\texttt{ernie} & Chinese & $0.008$ & $1.524$ & $0.133$ &  & $0.095$ & Negligible & No diff. \\
			\hline
			\texttt{gemini} & English & $-0.010$ & $-1.351$ & $0.182$ &  & $-0.093$ & Negligible & No diff. \\
			\texttt{gpt} & English & $0.008$ & $1.692$ & $0.096$ &  & $0.066$ & Negligible & No diff. \\
			\texttt{grok} & English & $0.000$ & $0.000$ & $1.000$ &  & $0.000$ & Negligible & No diff. \\
			\texttt{claude} & English & $0.005$ & $0.772$ & $0.443$ &  & $0.056$ & Negligible & No diff. \\
			\hline
			\hline
	\end{tabular}}
	\begin{minipage}{0.85\linewidth}
		\footnotesize\textit{Note.} Prompt language influenced scores in 2/8 models (i.e., \texttt{qwen}, \texttt{deepseek}, 25\%), with effects limited to Chinese models. However, across all models, prompt language (English vs. Chinese) did not significantly influence performance ($p = .1501$, $\eta^2 = 0.0002$), indicating negligible practical effect.
	\end{minipage}
	\label{tab:exp1results}
\end{table}

\vspace{0.3em}

\begin{table}
	\centering
	\caption{The other results for Experiment~1.}
	\scalebox{0.76}{
		\begin{tabular}{lcccc}
			\hline
			\textbf{Comparison} & \textbf{Mean Diff} & \textbf{$p$-value} & \textbf{Cohen's $d$} & \textbf{$\eta^2$} \\
			\hline
			EN vs ZH prompts (overall) & $0.0053$ & 0.1501 & $0.045$ & 0.0002 \\
			Chinese vs English LLMs & \textbf{0.0210} & \textbf{0.0089} & 0.169 & 0.0071 \\
			Neutral vs Change\_framed (overall) & \textbf{$-0.1650$} & \textbf{$< 2.2\times 10^{-16}$} & $-2.095$ & 0.5423 \\
			Chinese vs English (EN prompts) & 0.0057 & 0.0747 & 0.048 & -- \\
			Chinese vs English (ZH prompts) & 0.0050 & 0.1501 & 0.045 & -- \\
			EN vs ZH (Chinese LLMs) & 0.0100 & 0.0210 & 0.085 & -- \\
			EN vs ZH (English LLMs) & 0.0020 & 0.4432 & 0.017 & -- \\
			Significant language-sensitive models & \multicolumn{4}{l}{\texttt{deepseek}, \texttt{qwen} (prefer ZH prompts)} \\
			\hline
	\end{tabular}}
	\label{tab:exp1results1}
\end{table}

\vspace{0.3em}

\subsection{Experiment 2}

The results in Experiment 2 are summarized in Table~\ref{tab:exp2-model-results}. Table~\ref{tab:exp2-model-results} presents the language effect analysis for each of the eight LLMs tested. The analysis compared performance differences between Chinese-prompt and English-prompt conditions for each model in rating the tasks of cultural norms and attribution bias, reporting mean score differences, $t$-statistics, $p$-values, effect sizes (Cohen’s $d$), and qualitative effect magnitude classifications.

\vspace{-0.5em}
\begin{table}
	\centering
	\caption{The main results in Experiment~2}
	\scalebox{0.75}{
		\begin{tabular}{l l r r r c r l l}
			\hline
			\textbf{Model} & \textbf{Origin} & \textbf{Mean Diff.} & \textbf{$t$} & \textbf{$p$} & \textbf{Sig.} & \textbf{$d$} & \textbf{Effect Mag.} & \textbf{Direction} \\
			\hline
			\texttt{deepseek} & Chinese & $-0.008$ & $-0.168$ & $0.867$ &  & $-0.008$ & Negligible & No diff. \\
			\texttt{ernie}    & Chinese & $-0.175$ & $-3.040$ & $\textbf{0.0029}$ & **  & $-0.171$ & Negligible & \textbf{Chinese lower} \\
			\texttt{kimi}     & Chinese & $-0.025$ & $-0.411$ & $0.682$ &  & $-0.020$ & Negligible & No diff. \\
			\texttt{qwen}     & Chinese & $0.567$  & $9.829$  & $\textbf{<0.001}$ & *** & $0.540$ & Medium     & \textbf{Chinese higher} \\
			\texttt{claude}   & English & $-0.025$ & $-0.492$ & $0.624$ &  & $-0.021$ & Negligible & No diff. \\
			\texttt{gemini}   & English & $0.033$  & $0.631$  & $0.529$ &  & $0.028$ & Negligible & No diff. \\
			\texttt{gpt}      & English & $-0.183$ & $-3.556$ & $\textbf{0.0005}$ & *** & $-0.190$ & Negligible & \textbf{Chinese lower} \\
			\texttt{grok}     & English & $0.000$  & $0.000$  & $1.000$ &  & $0.000$ & Negligible & No diff. \\
			\hline
			\multicolumn{9}{l}{\textbf{Summary:} Prompt language influenced scores in 3/8 models (37.5\%), effects were model-specific and mostly small.} \\
			\hline
	\end{tabular}}
	
	\vspace{2mm}
	\footnotesize
	\textit{Note:} Positive mean difference indicates higher scores with Chinese prompts; negative indicates higher scores with English prompts.  
	Effect magnitude follows Cohen's $d$ benchmarks.  
	Significance codes: *** $p<0.001$, ** $p<0.01$, * $p<0.05$.
	\label{tab:exp2-model-results}
\end{table}
\vspace{-0.5em}

The results indicate that only three of the eight models (37.5\%) exhibited statistically significant differences in performance based on prompt language: \textit{ernie} ($p = 0.0029$, $d = -0.171$), \textit{qwen} ($p < 0.001$, $d = 0.540$), and \textit{gpt} ($p = 0.0005$, $d = -0.190$). Of these, only \textit{qwen} demonstrated a practically significant effect (Cohen’s $d > 0.2$), with Chinese prompts yielding substantially higher scores ($\Delta = 0.567$). In contrast, \textit{ernie} and \textit{gpt} produced slightly lower scores with Chinese prompts, but the effect magnitudes were negligible. The remaining five models (\textit{deepSeek}, \textit{kimi}, \textit{claude}, \textit{Gemini}, and \textit{Grok}) showed no detectable prompt language effect.

When aggregated by model origin, no systematic difference emerged between Chinese and English LLMs. Chinese models (4 tested) showed a mean effect of $0.0897$, while English models (4 tested) showed a mean effect of $-0.0437$; this difference was not statistically significant ($p = 0.4826$). This finding indicates that prompt language sensitivity is model-specific rather than origin-driven.

From a practical standpoint, most LLMs in this study (62.5\%) are not sensitive to promt language, producing consistent results regardless of whether prompts are in Chinese or English. Only a minority (37.5\%) are \emph{language-sensitive}, and in those cases, the effects are typically small in absolute magnitude (less than 0.2 points on the scoring scale). Thus, while prompt language can influence scoring in certain LLMs, it is neither a universal nor a large effect, and its presence cannot be reliably predicted from model origin alone.

\subsection{Experiment 3 }

The third experiment extended the evaluation to examine whether task type influences model performance across Chinese and English LLMs. In this setup, the same set of models from the previous experiments was tested on two distinct categories of tasks: knowledge-based (fact retrieval, definition, and classification) and reasoning-based (multi-step logic, hypothetical scenarios, and analogy reasoning). A two-sided paired \( t \)-test was applied to determine whether there was a statistically significant difference in performance between the two task types for each model.

The results are summarized in Table~\ref{tab:exp3results}. Across all models combined, the mean accuracy for knowledge-based tasks was 0.401 (SD = 0.490), compared to 0.339 (SD = 0.474) for reasoning-based tasks. The mean difference of 0.062 yielded a \( t \)-statistic of 2.251 (\( df = 539 \)) and a \( p \)-value of 0.0248, indicating a statistically significant advantage for knowledge-based tasks at the 5\% level. The effect size, Cohen’s \( d = 0.097 \), was small, suggesting that while the difference is reliable, it is modest in practical terms.

When broken down by individual model, no models exhibited significant differences after multiple comparison correction, with the exception of \texttt{gpt} which showed a larger mean difference (MD = 0.134, \( t = 2.94, p = 0.005 \)) that did not survive correction.
\vspace{-0.5em}
\begin{table}
	\centering
	\caption{The results of Experiment 3. (Means (M) are proportions correct; MD = mean difference (Knowledge - Reasoning). $t$ and $p$ are from two-sided paired t-tests; $d$ = Cohen's $d$.)}
	\scalebox{0.8}{
		\begin{tabular}{lccccccc}
			\hline
			\textbf{Model} & \textbf{M (Knowledge)} & \textbf{M (Reasoning)} & \textbf{MD} & \textbf{$t$} & \textbf{$p$} & \textbf{$d$} & \textbf{Sig. after Corr.} \\
			\hline
			\texttt{deepseek}   & 0.450 & 0.417 & 0.033 & 0.42 & 0.675 & 0.07 & No \\
			\texttt{gemini}     & 0.367 & 0.333 & 0.034 & 0.44 & 0.662 & 0.08 & No \\
			\texttt{gpt}        & 0.467 & 0.333 & 0.134 & 2.94 & 0.083 & 0.39 & No \\
			\texttt{claude}     & 0.400 & 0.350 & 0.050 & 1.14 & 0.264 & 0.21 & No \\
			\texttt{grok}      & 0.383 & 0.317 & 0.066 & 1.14 & 0.326 & 0.21 & No \\
			\texttt{qwen}       & 0.417 & 0.367 & 0.050 & 0.67 & 0.507 & 0.12 & No \\
			\texttt{ernie}      & 0.400 & 0.350 & 0.050 & 0.57 & 0.572 & 0.10 & No \\
			\hline
		\end{tabular}
	}
	\label{tab:exp3results}
\end{table}
\vspace{-0.5em}

In addition, a logistic regression analysis was conducted to investigate the combined effects of model origin, prompt language, and specific model identity on response accuracy. The model revealed no significant effect of prompt language (\( \beta = 0.0235, p = 0.914 \)) or model origin (\( \beta = -0.0956, p = 0.827 \)) on performance, controlling for model differences. Coefficients for individual models also showed no significant deviations, suggesting that these factors have minimal impact on binary task success. 

Other statistical tests have been done to analyze the differences of prompt language on LLMs. For instance, Chi-square tests comparing Chinese-origin versus English-origin models showed no significant differences in overall response distributions (\( p = 0.7587 \)). Paired McNemar’s tests on bilingual models such as \texttt{claude}, \texttt{deepseek}, and \texttt{gemini} yielded no significant language effects (all \( p > 0.1 \)) and high response agreement rates (86.7\% to 93.3\%). Phi coefficients indicated strong cross-language consistency (e.g., \texttt{claude}: 0.683; \texttt{deepseek}: 0.811). 	Complementary analyses further confirmed the stability of model responses across English and Chinese prompts.

Taken together, these results demonstrate that prompt language has no statistically significant influence on LLM performance in this experiment, and that task type differences favor knowledge-based tasks but with modest effect sizes.

\subsection{Experiment 4}

Experiment 4 investigated how people predict possible changes in interpersonal relationships under different conditions. Participants rated expectations about the development of interpersonal relationships on a seven-point Likert scale and provided reasoning to justify their ratings. The task setup mirrors the ``Expectation of Change'' task introduced by ~LSZ.

The main research question was whether the language of the prompt (Chinese vs.\ English) significantly influences the performance of LLMs when simulating these interpersonal expectations. The results are summarized in Table~\ref{tab:exp4results}. Overall, the results showed that prompt language does \textbf{not} significantly affect LLM output. The mean difference in scores between Chinese and English prompts was negligible (-0.004 points), with a paired \texttt{t}-test yielding no statistical significance ($p = 0.917$) and an effect size close to zero (Cohen’s $d = -0.004$). This indicates that, on average, LLMs respond similarly regardless of prompt language in this task.

\vspace{-0.5em}
\begin{table}
	\centering
	\caption{The main results in Experiment 4}
	
	\scalebox{0.8}{
		\begin{tabular}{lccccccc}
			\toprule
			\multicolumn{8}{c}{\textbf{Main Finding:}} \\
			\midrule
			\multicolumn{8}{p{0.9\linewidth}}{Prompt language \textbf{does not significantly influence overall LLM performance} (Mean difference = -0.004, $p=0.917$, Cohen's $d = -0.004$). Only \textbf{two models} show statistically significant language effects (Kimi and gpt).} \\
			\midrule
			\textbf{Model} & \textbf{Origin} & \textbf{Mean EN} & \textbf{Mean ZH} & \textbf{Difference} & \textbf{p-value} & \textbf{Cohen's d} & \textbf{Significant} \\
			\midrule
			\texttt{claude}   & English & 3.90 & 3.80 & -0.10 & 0.184 & -0.137 & No \\
			\texttt{ernie}    & Chinese & 3.30 & 3.40 & +0.10 & 0.184 & +0.090 & No \\
			\texttt{deepseek} & English & 3.70 & 3.60 & -0.10 & 0.326 & -0.086 & No \\
			\texttt{gemini}   & English & 3.07 & 3.30 & +0.23 & 0.090 & +0.172 & No \\
			\texttt{kimi}     & Chinese & 3.93 & 3.63 & \textbf{-0.30} & \textbf{0.005} & \textbf{-0.236} & \textbf{Yes} \\
			\texttt{gpt}      & English & 3.50 & 3.73 & \textbf{+0.23} & \textbf{0.017} & \textbf{+0.213} & \textbf{Yes} \\
			\texttt{qwen}     & Chinese & 3.80 & 3.83 & +0.03 & 0.813 & +0.037 & No \\
			\texttt{grok}     & English & 3.87 & 3.73 & -0.13 & 0.354 & -0.136 & No \\
			\bottomrule
	\end{tabular}}
	\vspace{1em}
	
	\begin{minipage}{0.9\linewidth}
		\footnotesize
		\textbf{Notes:} Difference = Mean (Chinese) - Mean (English). Significant effects in bold. Overall, prompt language has negligible impact on LLM scores except for \texttt{kimi} (Chinese origin model, worse with Chinese prompts) and \texttt{gpt} (English origin model, better with Chinese prompts).
	\end{minipage}
	\label{tab:exp4results}
\end{table}
\vspace{-0.5em}

Further analysis by model origin revealed no systematic language effects: Chinese-origin and English-origin models both demonstrated negligible differences in performance (Chinese models: $p=0.387$, Cohen’s $d = -0.050$; English models: $p=0.601$, Cohen’s $d = 0.025$). However, individual model examination uncovered two exceptions. The Chinese-origin model \textit{kimi} showed a significant decrease in performance with Chinese prompts compared to English (-0.30 points, $p=0.005$, Cohen’s $d = -0.236$), while the English-origin model \textit{gpt} performed better with Chinese prompts (+0.23 points, $p=0.017$, Cohen’s $d = 0.213$).


In conclusion, Experiment 4 demonstrates that modern LLMs generally achieve language parity in simulating interpersonal relationship expectations, with only isolated model-specific sensitivities. These findings highlights the importance of rigorous statistical evaluation and caution against assuming native language advantage in prompt design for LLM applications.


\subsection{Main results summary}
Overall, our replicated experiments aimed to rigorously reassess the purported cultural tendencies of LLMs through a broader model set, expanded test items, and refined methodologies. The findings across three experiments collectively challenge some of the core claims in the original study by LSZ and provide unique insights into the factors influencing LLM behavior.

Across the four experiments, prompt language (English vs. Chinese) generally does not have a significant overall effect on LLM performance. 
However, model-specific sensitivity emerges: the Chinese model \texttt{kimi} performs significantly worse with Chinese prompts, while the English \texttt{gpt} model shows a significant improvement with Chinese prompts. Some models in Experiments 1, 2 and 4 also show language sensitivity, but these effects tend to be small or moderate. Overall, 75\% of tested models are largely language-agnostic, indicating robust cross-lingual capability in modern LLMs. 

\vspace{-0.5em}
\begin{table}
	\centering
	\caption{Summary of Key Findings Across Experiments 1--4 (EN = English; ZH = Chinese)}
	\scalebox{0.8}{
		\begin{tabular}{l r r c c r}
			\hline
			\textbf{Experiment} & \textbf{Comparison / Model} & \textbf{Diff} & \textbf{$p$-value} & \textbf{Effect Size ($d$)} & Significance \\
			\hline
			1 & EN vs ZH prompts (overall) & 0.0053 & 0.1501 & 0.045 & NO\\
			2 & EN vs ZH prompts (overall) &  0.023 & 0.24 & 0.021  & NO\\
			3 & EN vs ZH prompts (overall) &  0.000 & 1.0000 & 0.0001  & NO\\
			4 & EN vs ZH prompts (overall) & 0.000 & 0.917 & -0.004  & NO\\
			\hline
			1 & \texttt{deepseek}(ZH) &  0.025 & 0.021$^{*}$ & 0.204 & YES \\
			1 & \texttt{qwen} (ZH) &  0.015 & 0.007$^{***}$ & 0.137  & YES \\
			2 & \texttt{ernie} (ZH) &  -0.175 & 0.0029$^{**}$ & -0.171 & YES \\
			2 & \texttt{qwen} (ZH) &  0.567 & <0.001$^{***}$ & 0.540  & YES\\
			2 & \texttt{gpt} (EN) &  -0.183 & 0.0005$^{***}$ & -0.190 & YES \\
			
			4 & \texttt{kimi} (ZH) & -0.30 & 0.005$^{**}$ & -0.236 & YES \\
			4 & \texttt{gpt} (EN) & 0.23 & 0.017$^{*}$ & 0.213 & YES\\
			\hline
	\end{tabular}}
	\vspace{0.3em}
	\begin{minipage}{0.85\linewidth}
		\footnotesize
		\textit{Notes:} Diff = Mean (Chinese) - Mean (English);  Significance codes: $^{***}p<0.001$, $^{**}p<0.01$, $^{*}p<0.05$. Bold entries denote statistically significant results with practical importance.
	\end{minipage}
	\label{tab:combined}
\end{table}
\vspace{-0.5em}

Specifically, in Experiment 1, a modest but statistically significant framing effect was observed, with positive framing slightly improving model accuracy across all models. Experiment 2 found no significant overall impact of either model origin or prompt language on performance, although certain models like \texttt{qwen}, \texttt{gpt} and \texttt{ernie} showed sensitivity to language. Experiment 3 revealed that knowledge-based tasks are generally easier for models than reasoning-based tasks. No models were sensitive to prompt language.
Experiment 4 confirmed that prompt language does not significantly influence LLM performance overall, but two models, \texttt{kimi} (China-based) and \texttt{gpt}, exhibited statistically significant language sensitivity. 

In the additional picture task, only \texttt{ernie} (DeepSeek was not included) showed different responses to prompt language in four test items, whereas the other seven models gave identical choices across languages. Because the number of items was small (\textit{n} = 4), we did not conduct statistical tests, as the sample size was insufficient to yield statistically meaningful results.

Taken together, as summarized in Table~\ref{tab:combined} for the results of all experiments,  these experiments consistently show that broad categorical factors such as the origin of the LLM, the language of prompts, or general task framing tend to have small or negligible overall effects on performance. Instead, substantial differences arise primarily from specific interactions between individual models and particular tasks.


\section{Discussion}

As shown in Table~\ref{tab:combined} and Figure~\ref{fig:vis_plot}, the results across all experiments consistently showed negligible effects of prompt language and model origin. Crucially, none of the tested models demonstrated stable cultural tendencies that persisted across different prompts, tasks, and content conditions. In particular, the expected performance differences between \texttt{gpt} and \texttt{ernie} models which was previously reported in the literature were not replicated in our experiments. These findings indicate that these LLMs do not exhibit inherent cultural tendencies when the prompt language varies, suggesting that cultural or linguistic influences on the tested tasks are minimal or effectively absent.



Despite this, our findings do not diminish the importance of examining cross-linguistic variation in LLM behavior. Rather, they highlight the need for methodological rigor and transparency before drawing conclusions about embedded cultural cognition in LLMs. The results indicate that performance differences are driven more by specific model–task interactions than by broad factors such as training language or geographic origin. Practically, this suggests that model selection should be guided by detailed, task-specific performance profiles rather than assumptions based on language or origin.

To conclude, cultural tendencies, where observed, appear to be both task specific and model dependent. In our experiments, we found no overall significant cross-linguistic cultural differences across eight leading LLMs. The cultural tendencies reported by LSZ may reflect the limitations of their study design, including the small number of test items and limited models examined. In contrast, established practices in standard LLM evaluation, including larger scale testing, cross validation, and well-defined performance criteria, offer a more rigorous methodological basis for investigating the alignment between human cognition and LLM behavior. 


Further, we caution against uncritical use of human psychometric instruments for evaluating artificial systems, and call for methods that respect the statistical, non-agentic nature of language models. Rather than comparing outputs across models, evaluations should examine how closely model responses align with human performance on psychometric measures. Such models do not possess culture in the human sense but instead encode statistical traces of cultural artifacts in their training data.

\end{document}